\documentclass[10pt,twocolumn,letterpaper]{article}

\usepackage[pagenumbers]{iccv} 

\usepackage[export]{adjustbox}  %
\usepackage{makecell}  %

\newcommand{\OURS}{Pixel3DMM}

\usepackage{booktabs} %
\usepackage{comment}
\usepackage{tabularx}
\usepackage{multirow}
\usepackage{caption}
\usepackage{subcaption}
\usepackage{dsfont}

\definecolor{iccvblue}{rgb}{0.21,0.49,0.74}
\usepackage[pagebackref,breaklinks,colorlinks,allcolors=iccvblue]{hyperref}

\title{
Pixel3DMM: Versatile Screen-Space Priors for \\ Single-Image 3D Face Reconstruction
}

\author{
Simon Giebenhain$^1$ \quad
Tobias Kirschstein$^1$ \quad
Martin Rünz$^2$ \\
Lourdes Agapito$^3$ \quad
Matthias Nie{\ss}ner$^1$ \vspace{0.2cm}\\
$^1$Technical University of Munich  \qquad $^2$Synthesia \qquad $^3$University College London
}

\begin{document}

\twocolumn[{
\renewcommand\twocolumn[1][]{#1}
\maketitle
\thispagestyle{empty}
\begin{center}
  \newcommand{\teaserwidth}{\textwidth}
   \vspace{-0.8cm}

  \centerline{

    \includegraphics[width=\teaserwidth]{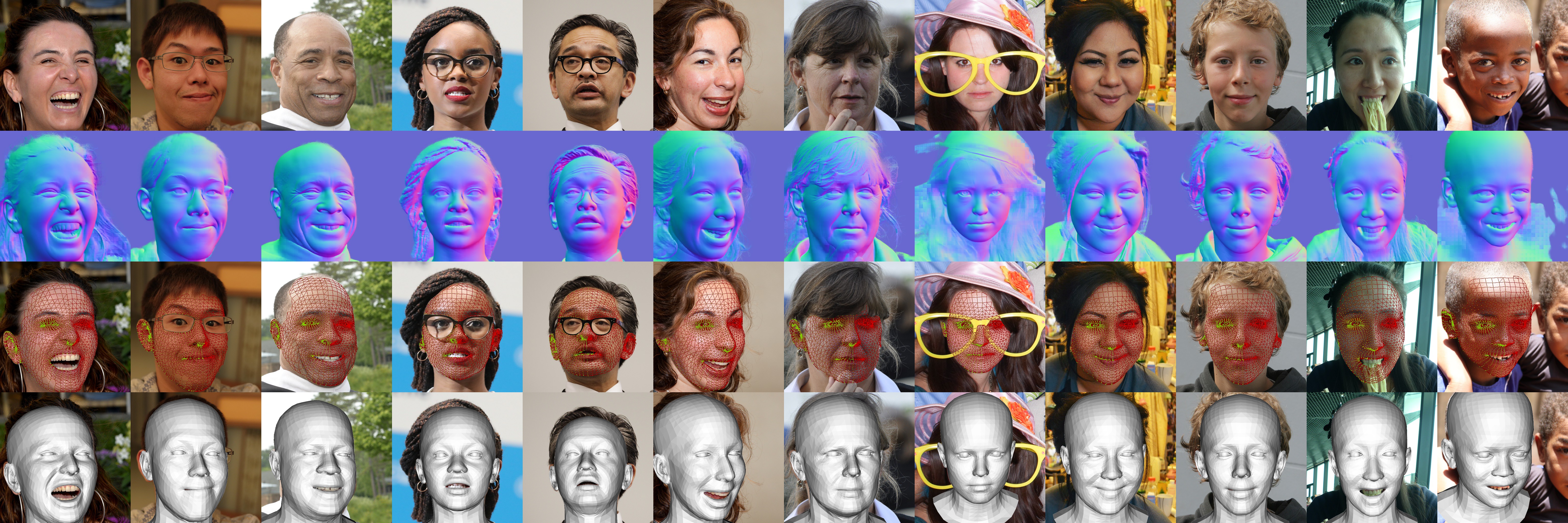}
    }
     \vspace{-0.2cm}
    \captionof{figure}{
    We present \OURS, a set of two ViTs~\cite{dosovitskiy2020vit}, which are tailored to predict per-pixel surface normals and uv-coordinates. 
    Here, we demonstrate the fidelity and robustness of our prior networks on examples from the FFHQ~\cite{karras2019ffhq} dataset. From top to bottom we show input RGB, predicted surface normals, 2D vertices extracted from the uv-coordinate prediction, and our FLAME fitting results.
    }
  \label{fig:teaser}
      \vspace{-0.2cm}

 \end{center}
}]

\maketitle

{\let\thefootnote\relax\footnotetext{\scriptsize{Project Page: \url{https://simongiebenhain.github.io/pixel3dmm}}}}

\begin{abstract}
We address the 3D reconstruction of human faces from a single RGB image.
To this end, we propose \OURS, a set of highly-generalized vision transformers which predict per-pixel geometric cues in order to constrain the optimization of a 3D morphable face model (3DMM). 
We exploit the latent features of the DINO foundation model, and introduce a tailored surface normal and uv-coordinate prediction head. 
We train our model by registering three high-quality 3D face datasets against the FLAME mesh topology, which results in a total of over 1,000 identities and 976K images.
For 3D face reconstruction, we propose a FLAME fitting opitmization that solves for the 3DMM parameters from the uv-coordinate and normal estimates.
To evaluate our method, we introduce a new benchmark for single-image face reconstruction, which features high diversity facial expressions, viewing angles, and ethnicities. 
Crucially, our benchmark is the first to evaluate both posed and neutral facial geometry.
Ultimately, our method outperforms the most competitive baselines by over 15\% in terms of geometric accuracy for posed facial expressions. %

\end{abstract}

\section{Introduction}
\label{sec:intro}

3D reconstruction of faces, tracking facial movements, and ultimately extracting expressions for animation tasks are fundamental problems in many domains such as computer games, movie production, telecommunication, and AR/VR applications. 
Recovering 3D head geometry from a single image is a particularly important task due to the vast amount of available image collections. %

Unfortunately, reconstructing faces from a single input image is also inherently under-constrained. Not only depth ambiguity renders this task challenging, but also ambiguities between albedo and lighting/shadow effects. In addition, properly disentangling identity and expression information -- which is critical for many downstream applications -- makes the problem difficult.
Finally, occlusions and unobserved facial regions further complicate the problem in real application scenarios, thus highlighting the need for strong data priors. %

A typical approach to address single-image face reconstruction is to exploit 3D parametric head models (3DMMs)~\cite{blanz20233dmm, FLAME} which provide a comparatively low-dimensional parametric representation for the underlying 3D geometry.
Optimizing within a 3DMM's disentangled parameter space heavily constrains the search space with built-in assumptions about plausible facial structure and expressions, and allows to extract disentangled identity and expression information.
Nonetheless, despite relying on 3DMMs, many ambiguities remain and their simplifying assumptions about our world often cannot explain the complexity of an observed RGB signal.
This necessitates additional priors in order to obtain compelling fitting results such as sparse~\cite{landmarks} and dense~\cite{cao20133d, wood2022denselandmarks} facial landmarks, or UV coodinate predictions~\cite{taubner2024flowface}

In recent years, we have also seen significant progress in feed-forward 3DMM regressors~\cite{Sanyal2019now, feng2021learning, danvevcek2022emoca, SMIRK:CVPR:2024, zielonka2022towards, zhang2023accurate}. However, it is complicated to extend feed-forward regressors, \eg to a multi-view or temporal domain, and, as we will show later, they fall behind optimization-based approaches on inputs with strong facial expressions.
Overall, accurate 3D face reconstruction from single images remains a challenging and highly relevant problem.

Therefore, we propose \OURS, a novel optimization-based 3D face reconstruction approach. %
Our main idea is to exploit and further develop broadly generalized and powerful foundation models to predict pixel-aligned geometric cues that effectively constrain the 3D state of an observed face. 
Given a single image at test time, we propose normal and uv-coordinate predictions as optimization constraints from which we fit a 3D FLAME model. 
Instead of a simple rendering loss of uv-coordinates, we then transfer the information into a 2D vertex loss, which offers a wider basin of attraction during optimization.
We argue that this strategy is superior to traditional photometric terms, or sparse landmarks, 
which often struggle with extreme view points and facial expressions.
In order to train our approach, we unify three recent, high-fidelity 3D face datasets \cite{giebenhain2023nphm,zhu2023facescape,martinez2024ava256} by registering them against the FLAME~\cite{FLAME} model.%
Our approach outperforms all available normal estimators for human faces in the NeRSemble~\cite{kirschstein2023nersemble} dataset.%

In order to advance the evaluation of single-image 3D face reconstruction methods, we further propose a new benchmark based on the multi-view video dataset NeRSemble~\cite{kirschstein2023nersemble}, which includes a wider variety of facial expressions than existing benchmarks~\cite{Sanyal2019now, zhu2023facescape, feng2018stirling, Chai2022REALY}.
Our benchmark is the first to allow for the simultaneous evaluation of posed and neutral facial geometry. 
This enables a more direct comparison of methods, especially regarding fitting fidelity and ability to disentangle expression and identity information. 
Finally, we show that compared to our strongest baselines, our approach improves the L2-Chamfer reconstructions loss by over 15\% for posed geometry, while slightly improving over neutral geometry predictions.

\medskip

\noindent
To summarize, our main contributions are as follows:
\begin{itemize}
    \item A new formulation to exploit foundation model features for 3D-related, pixel-aligned predictions, facilitating state-of-the-art normal estimations for human faces. 
    \item A novel 3D face reconstruction approach based on predicted uv-map correspondences and surface normals.
    \item A 3D face reconstruction benchmark and evaluation protocol from high-fidelity multi-view face captures.   
\end{itemize}
We plan to make the model, code, and our new benchmark publicly available to promote progress in single image 3D face reconstruction and encourage quantitative benchmarking on challenging facial expressions.

\begin{figure*}[htb!]
    \centering
    \includegraphics[width=0.99\textwidth]{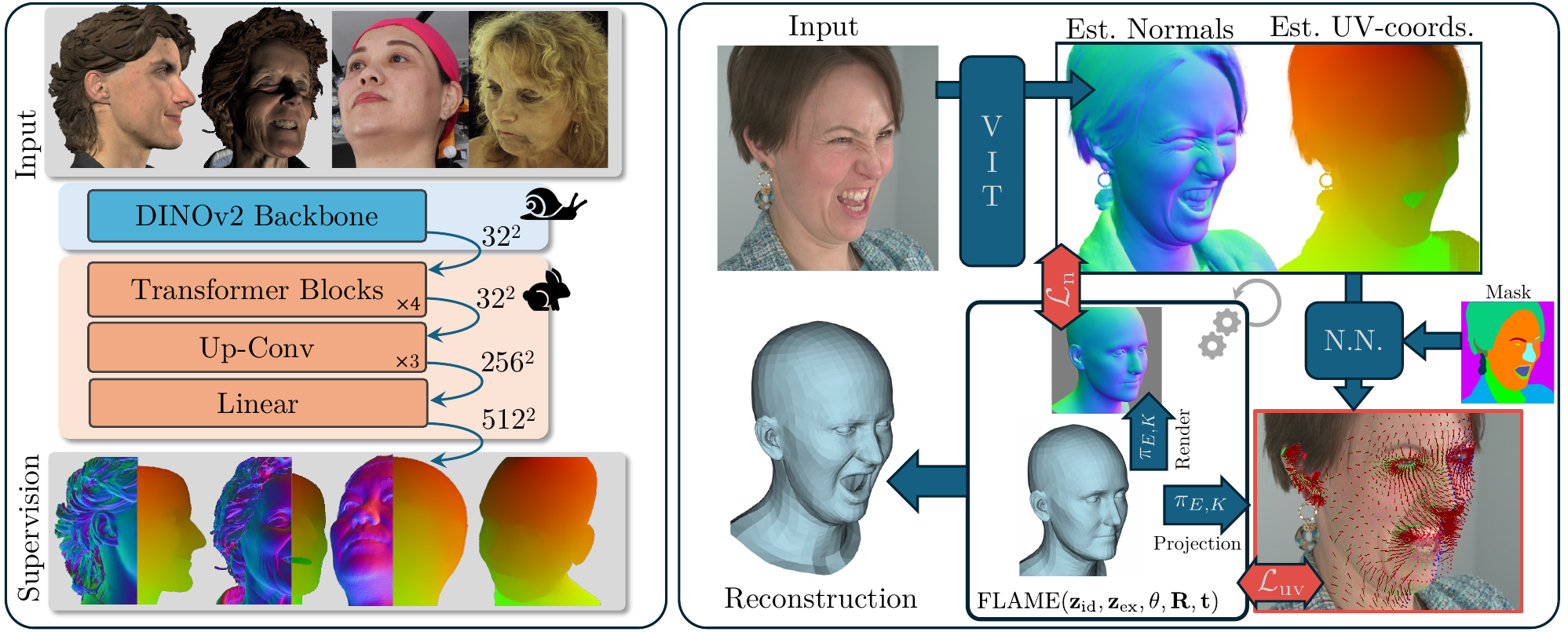}

    \caption{
    \textbf{Method Overview:} 
    {\OURS} consists of (a) learning pixel-aligned geometric priors (left) and (b) test-time optimization against predicted uv-coordinates and normals (right).
    On the left we illustrate our network architecture and examples from the training set.
    On the right we illustrate to process of finding per-vertex 2D locations using a nearest neighbor (N.N.) look up, and our loss terms.
    }
    \label{fig:method}
    \vspace{-0.3cm}

\end{figure*}

\section{Related Work}
\label{sec:related_work}

\noindent\textbf{Single-Image 3DMM Fitting~ }
Tracking morphable models from single images is a well-studied problem in the context of 3D face reconstruction and tracking. Early works~\cite{blanz1999morphable, paysan20093d, FLAME}, introduced statistical shape and texture priors to estimate 3D face geometry from 2D images. Such methods rely on photometric fitting and subsequent approaches improve modeling capabilities using learned implicit representations~\cite{lin2023single, giebenhain2024mononphm}. While some methods~\cite{thies2016face2face, grishchenko2020attention} favor a high tracking frame rate for real-time applications, others favor reconstruction accuracy~\cite{zielonka2022towards}. \\

\noindent\textbf{Facial Landmark Prediction~ }
Numerous reconstruction methods~\cite{FLAME, cao20133d} for faces rely on accurate landmark predictions, which are usually coupled with vertices of a template mesh. Pioneering work on detecting such landmarks already relies on statistical learning~\cite{cootes2001active} and more recent models exploit large datasets~\cite{wood2021fake, wayne2018lab} and neural networks to improve the performance~\cite{bulat2017far, bazarevsky2019blazeface}. MediaPipe~\cite{bazarevsky2019blazeface}, for instance, uses a convolutional network inspired by MobileNet~\cite{howard2017mobilenets}.

Another line of work focuses on densely aligning template mesh and 2D predictions. To achieve this FlowFace~\cite{taubner2024flowface} employs a vision-transformer backbone and iteratively refines the flow from UV to image space.\\

\noindent\textbf{3DMM Regression~ } 
DECA~\cite{feng2021learning} trains an encoder for 3DMM parameters that also outputs a displacement map for higher fidelity. An extension of this work is presented in EMOCA~\cite{danvevcek2022emoca}, which adds a head for facial expressions to the architecture and emphasises on the reconstruction of emotion-rich data.  SPECTRE~\cite{filntisis2022visual} too builds on top of DECA, but aims at temporal consistency and reconstructing lip motion truthfully. To improve the analysis-by-synthesis aspect of previous methods SMIRK~\cite{SMIRK:CVPR:2024} introduces a neural synthesis component, reducing the domain gap between real and rendered images.
Since the aforementioned methods don't assume 2D to 3D correspondences for training, it is easy to scale them to large datasets. As a downside, the lack of 3D information impedes accuracy and leaves depth ambiguity. In order to address this, MICA~\cite{zielonka2022towards} supervises directly on meshes and presents a 3D face dataset to do so. TokenFace~\cite{zhang2023accurate} is a transformer-based hybrid method that can be trained on both 2D and paired 3D data. \\

\noindent\textbf{Face Reconstruction Benchmarks~ }
Benchmarks with high-quality ground truth reconstructions are necessary to compare methods reliably. The Stirling~\cite{feng2018stirling} dataset contains 2000 images of
135 subjects. %
Unfortunately, ground truth reconstructions are only available for neutral poses in this dataset. Similarly, the NoW~\cite{Sanyal2019now} benchmark provides reconstructions only in the neutral expression. It has 2054 images of 100 subjects and 3D models recorded with an active stereo 3dMD system.
Both the FaceScape~\cite{zhu2023facescape} and the REALY~\cite{Chai2022REALY} dataset contain posed scans. While the former has 10 identities, the latter has 100 subjects. Neither of these two benchmarks measures disentanglement by additionally evaluating against neutral geometry.

\section{\OURS}
\label{sec:pixel3dmm}

In this work we address the challenges of single-image face reconstruction by learning powerful priors of pixel-aligned geometric cues. In particular we train two vision transformer networks, which predict uv-coordinates and surface normals against which we fit FLAME~\cite{FLAME} parameters at inference time. In \cref{sec:piror_learning} we describe our  {\OURS} networks, our data acquisition, and how we train them for accurate surface normal and uv-coordinate prediction.
Afterwards, in~\cref{sec:fitting}, we elaborate on our single-image fitting approach, which is purely based on our surface normal and uv-coordinate predictions.

\subsection{Learning Pixel-Aligned Geometric Cues}
\label{sec:piror_learning}

Despite recently released high-quality 3D face datasets~\cite{zhu2023facescape,kirschstein2023nersemble,giebenhain2023nphm,martinez2024ava256}, such data is still relatively scarce, especially w.r.t. the number of different identities, ethnicities, age distribution and lighting variation.
We therefore take inspiration from recent achievements on fine-tuning foundational and large generative models to become experts on a constrained domain, \eg \cite{hu2022lowrank, ruiz2023dreambooth}.

In particular we train two expert networks
\begin{eqnarray}
    \mathcal{N}&:& \mathbb{R}^{512\times 512 \times 3} \rightarrow [-1, 1]^{512 \times 512 \times 3}\\
    \mathcal{U}&:& \mathbb{R}^{512\times 512 \times 3} \rightarrow [~0, 1]^{512 \times 512 \times 2}
\end{eqnarray}
which, given a single input image $I$, predict surface normals $\mathcal{N}(I)$ and uv-space coordinates $\mathcal{U}(I)$, respectively.

\subsubsection{Network Architecture}

We build {\OURS} on top of the foundational features from a pre-trained DINOv2~\cite{oquab2023dinov2} backbone. As depicted in \cref{fig:method}, we extend the ViT architecture using a simple prediction head. 
It consists of four additional transformer blocks, three up-convolutions which lift the feature map resolution from $32 \time 32$ to $256\times256$. Finally, we use a single linear layer to increase the feature dimensionality and unpatchify the predictions to $512\times512\times c$, where $c\in\{3, 2\}$ for normals and uv-coordinate prediction tasks, respectively.

\subsubsection{Data Preparation}
To train our networks, we opt for three recent, high-quality 3D face datasets: NPHM~\cite{giebenhain2023nphm}, FaceScape~\cite{zhu2023facescape}, and Ava256~\cite{martinez2024ava256}. We follow the non-rigid registration procedure from NPHM, to obtain the same kind of high-quality registrations in FLAME topology for FaceScape and Ava256.

\cref{fig:method} shows pairs of input views with the associated supervision signal for surface normals and uv-coordinates. Since Ava256 does not provide high-fidelity geometry, we exclusively use it to supervise our UV-network $\mathcal{U}$.
Since the NPHM dataset only consists of textured meshes, we render 40 random views randomly distributed on the frontal hemisphere using randomized intrinsics and camera distances. Addtionally, we randomly sample lighting conditions (using point lights) and material parameters for each rendering.

\paragraph{Dataset Numbers}
In total, our dataset comprises 470 identities from NPHM in 23 expression and 40 renderings each (376K rgb, normal and uv images in total). For FaceScape we use 350 subjects, observed under 20 different expressions and 50 cameras each (350K rgb, normal and uv images in total). Since Ava256 is a video dataset, we leverage furthest point sampling to select the 50 most diverse expressions per person. For each person we choose a random subset of 20 cameras (250K rgb and uv images in total).

\paragraph{Diffsion-based Lighting Variations}

Since FaceScape and Ava256 are both studio datasets, which are captured at rather homogeneous lighting conditions, we leverage IC-Light~\cite{IC-Light}, an image conditioned diffusion model~\cite{rombach2022stablediffusion}, which alters the lighting condition based on a text prompt or background image.

\subsubsection{Training}

We train our models $\mathcal{M}\in \{\mathcal{N}, \mathcal{U} \}$ using a straight forward image translation formulation

\begin{equation}
    \underset{\Psi_{\mathcal{M}}}{\mathrm{argmin}} \sum_{k\in \mathcal{D}}\sum_{p \in M^k} \Vert f(I^k)_p - Y^k_p\Vert_2,
\end{equation}

where $\Psi_{\mathcal{M}}$ denotes the network's parameters, $k\in\mathcal{D}$ is a sample from our dataset, $I^k$ and $Y^k$ are input rgb and target images, respectively, and $p\in M^k$ are all pixels in the associated foreground mask.

Note, that instead of freezing the parameters of our {DINOv2} backbone altogether, we set their learning rate ten times lower, in order to encourage prior preservation but enable stronger domain adoption.

Compared to Sapiens~\cite{khirodkar2024sapiens}, a recent state-of-the-art foundation model for human bodies, training our models is cheap and can be realized using 2 GPUs and training for 3 days. Additionally, we highlight the fact that all data is publically available. The relatively low computational burden and data accessibility, will hopefully inspire more research to follow in a similar direction. 
Finally, note that uv-coordinates are an abstract concept, without a strong correlation to rgb data, requiring a more global understanding of the input. 
Therefore, we demonstrate that the available data is enough to achieve generalization on complicated, semi-global prediction tasks.

\subsection{Single-Image FLAME\cite{FLAME} Fitting}
\label{sec:fitting}

Given a single image $I$, we leverage our prior networks to obtain predicted surface normals $\mathcal{N}(I)$ and uv-coordinates $\mathcal{U}(I)$.
Using these predictions we aim to recover 3DMM parameters. In particular, we optimize for FLAME~\cite{FLAME} identity, expression, and jaw parameters, as well as, camera rotation, translation, focal length and principal point:

\begin{eqnarray}
    \Omega_{\text{FLAME}} &=& \{\mathbf{z}_{\text{id}} \in \mathbb{R}^{300}, 
                \mathbf{z}_{\text{ex}} \in \mathbb{R}^{100},
                \theta \in \mathcal{SO}(3) \} \\
    \Omega_{\text{cam}}\!&=&\!\{
                \mathbf{R}\!\in\!\mathcal{SO}(3),\!
                \mathbf{t}\!\in\!\mathbb{R}^{3},\!
                \mathbf{fl}\!\in\!\mathbb{R}^+,\!
                \mathbf{pp}\!\in\!\mathbb{R}^2\!
                \}.
\end{eqnarray}

\subsubsection{2D Vertex Loss}

Using the estimated uv-coordinates $\mathcal{U}(I)$, we aim to extract the 2d location $p^*_v$ for each visible vertex $v \in V$ of the FLAME mesh. 
To this end we first run a facial segmentation network~\cite{zheng2022farl}, in order to mask out the background, eyeballs and mouth interior. 
Then we find correspondences for each vertex $v\in V$ using a nearest neighbor lookup into $\mathcal{U}(I)$.
To be more specific let $T^{\text{uv}}_v \in [0, 1]^2$ denote the uv-coordinate of $v$ in the template mesh $T$. Then we find the pixel location 
\begin{equation}
   p^*_v =  \underset{p \in P}{\mathrm{argmin}} \Vert T^{\text{uv}}_v - \mathcal{U}(I)_p\Vert
\end{equation}
as the pixel with the closest uv prediction. Finally, we define 

\begin{equation}
    \mathcal{L}_{\text{uv}} = \sum_{v \in V} \mathds{1}_{\Vert T^{\text{uv}}_v - \mathcal{U}(I)_p\Vert < \delta_{\text{uv}} } \cdot \vert p^*_v - \pi(v) \vert 
\end{equation}

to be our 2d vertex loss, where $\mathds{1}$ denotes the indicator function which masks out vertices with a nearest neightbor distance larger than $\delta_{uv}$.
$V=\operatorname{FLAME}\left( \Omega_{\text{FLAME}}\right)$ is the current estimate of the FLAME parametric model, and $\pi$ denotes the projection implied by the current estimate of the camera parameters $\Omega_{\text{cam}}$.

\subsubsection{Optimization}
\label{sec:opt}

Next to the 2d vertex loss $\mathcal{L}_{uv}$, we include the normal loss $\mathcal{L}_{n} = \vert \mathcal{N}(I) - \texttt{render}_n(V)\vert$, where $\texttt{render}_n$ denotes a rendering of surface normals of the FLAME mesh. 
 The regularization term $\mathcal{R} = \lambda_{\text{id}}\Vert \mathbf{z}_{\text{id}} - \textbf{z}_{\text{id}}^{\text{MICA}}\Vert_2^2 + \lambda_{\text{ex}}\Vert \mathbf{z}_{\text{ex}}\Vert^2_2$ completes our overall energy term

\begin{equation}
    E = \lambda_{uv}\mathcal{L}_{uv} + \lambda_n \mathcal{L}_{\text{n}} + \mathcal{R}.
    \label{eq:energy}
\end{equation}

Here $\textbf{z}_{\text{id}}^{\text{MICA}}$ denotes MICA's~\cite{zielonka2022towards} identity prediction.

\subsection{Monocular Video Tracking}
\label{sec:tracking}

Next to the single-image scenario, tracking faces in monocular videos is a fundamental task in computer vision. To address this problem, we simply extend our optimization strategy from \cref{sec:opt} globally over all images in a video sequence $\{I_t\}_{t=1}^T$. 
Using our prior networks, we first obtain normal predictions $\{\mathcal{N}(I_t)\}$ and uv-predictions $\{\mathcal{U}(I_t)\}$
After obtaining an initial estimate for $\Omega^{(0)}_{\text{FLAME}}$ and $\Omega^{(0)}_{\text{cam}}$ on the first frame by optimizing for \cref{eq:energy}, we freeze $\mathbf{z}_{\text{id}}$, $\mathbf{fl}$ and $\mathbf{pp}$. We then sequentially optimize for all remaining attributes in $\Omega^{(t)}_{\text{FLAME}}$ and $\Omega^{(t)}_{\text{cam}}$. 
Using the results from the sequential optimization pass as initialization, we extend \cref{eq:energy} to a batched version including a random sample of $B=min(T, 16)$ frames. 
Note, that the parameters $\mathbf{z}_{\text{id}}$, $\mathbf{fl}$ and $\mathbf{pp}$ are shared for all frames. In order to enforce smoothness across all per-frame optimization targets we add a smoothness term \begin{equation}
\mathcal{L}_{\text{smooth}}^{\Phi}\!=\!\frac{\lambda_\text{smooth}^{\Phi}}{2*B}\!\sum_{t \in B}\!\Vert \Phi^{(t-1)}\!-\!\Phi^{(t)}\Vert_2^2\!+\!\Vert \Phi^{(t)}\!-\!\Phi^{(t+1)}\Vert_2^2
\end{equation}
to our optimization energy $E$, where we let $\Phi^{(t)} \in \{\mathbf{z}_{\text{ex}}^{(t)}, \theta^{(t)}, \mathbf{R}^{(t)}, \mathbf{t}^{(t)}\}$ denote any of the per-frame variables.

\begin{figure}[tb]
    \centering
    \begin{tabularx}{0.07\linewidth}{l}
        \rotatebox[origin=r]{90}{\parbox[c]{1.3cm}{\centering Ours}} \\
        \rotatebox[origin=r]{90}{\parbox[c]{1.3cm}{\centering FaceScape}} \\
        \rotatebox[origin=r]{90}{\parbox[c]{1.3cm}{\centering NoW}}
    \end{tabularx}%
    \includegraphics[width=0.93\linewidth, valign=c]{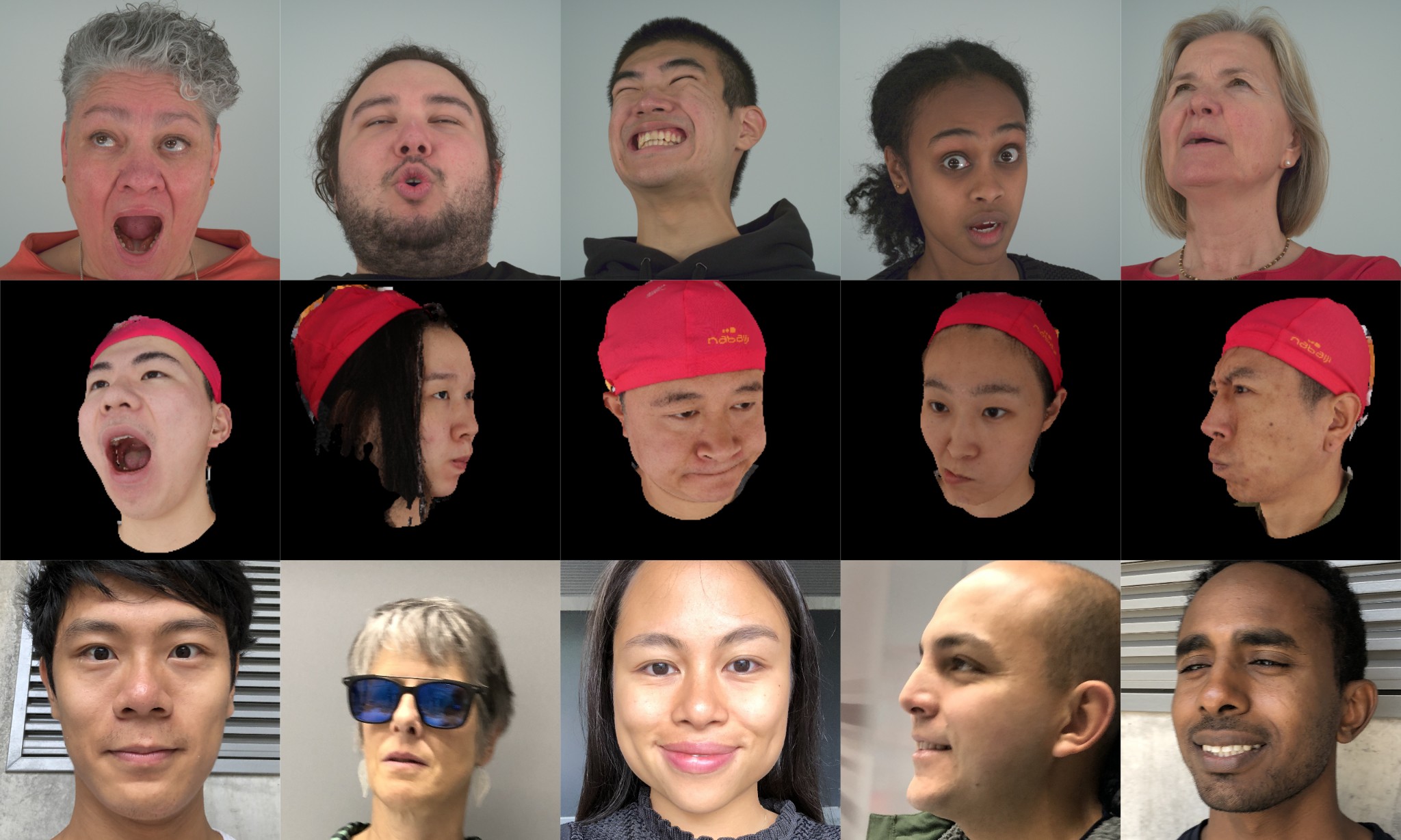}
    \caption{\textbf{3D Face Reconstruction Benchmark Analysis.} We show the 5 most diverse images from each benchmark dataset, as measured by the expression codes of EMOCA~\cite{danvevcek2022emoca}. Our benchmark covers a richer diversity of facial expressions.}
    \label{fig:benchmark_analysis}
    \vspace{-0.3cm}

\end{figure}
\begin{table}[htb]
    \centering
    \setlength{\tabcolsep}{3.5pt}
    \small
    \begin{tabular}{lrrrrrr}
    \toprule
         & \small{Year} & \small{neutr.} & \makecell{\small{expr.} \\ \small{div.}} & \makecell{\small{view} \\ \small{div.}} & \small{\#pers.} & \small{\#Scans} \\
    \midrule
        Stirling~\cite{feng2018stirling}
            & 2013 & \checkmark & & \checkmark & \textbf{133} & 133 \\
        REALY~\cite{Chai2022REALY}
            & 2015 &  &  & & 100 & 100 \\
        NoW~\cite{Sanyal2019now}
            & 2019 & \checkmark & & \checkmark & 80 & 80 \\
        FaceScape~\cite{zhu2023facescape}
            & 2020 & & \checkmark & \checkmark & 20 & 20 \\
        Ours 
            & 2023 & \checkmark & \checkmark & \checkmark & 21 & \textbf{441} \\
    \bottomrule
    \end{tabular}
    \caption{\textbf{Comparison of 3D Face Reconstruction Benchmarks.} We compare data capture year, whether the benchmark evaluates disentanglement by predicting a neutral mesh from a posed image (neutr.), expression diversity (div. expr.), viewpoint diversity (div. views), number of persons (\#pers.) and number of GT scans.}
    \label{tab:benchmark_comparison}
    \vspace{-0.3cm}

\end{table}
\section{3D Face Reconstruction Benchmark}
\label{sec:benchmark}

Human face geometry is complex due to the presence of thin structures, different textures and diverse shapes. Furthermore, humans can deform their facial geometry in a remarkable way, performing a wide range of expressions and emotions. Consequently, building a robust 3D face reconstruction pipeline that covers all potential states of a human face is a challenging endeavor. Several 3D face reconstruction benchmarks have been previously proposed to rank reconstruction methods in terms of quality and robustness.~\cref{tab:benchmark_comparison} shows a comparison of popular benchmarks. However, we find that most existing benchmarks rarely evaluate extreme facial expressions, an important aspect of human face geometry. This can be seen in~\cref{fig:benchmark_analysis} where we retrieve the 5 most expressive images from the recent FaceScape benchmark~\cite{zhu2023facescape} and the established NoW benchmark~\cite{Sanyal2019now}. We do this by running EMOCA~\cite{danvevcek2022emoca} on each image of the dataset, collecting the expression codes, and then performing furthest point sampling in EMOCA's expression space, starting from the expression with highest norm. We find that FaceScape only contains 20 different but relatively articulated expressions while the NoW benchmark is dominated by mostly neutral and smiling expressions.   
We therefore propose a new benchmark for 3D face reconstruction that is sourced from images of the recently published multi-view video dataset NeRSemble~\cite{kirschstein2023nersemble}. For 21 diverse identities, we select 20 distinct expressions via furthest point sampling of 3D landmarks for a total of 420 images. The corresponding ground truth 3D geometries are obtained by running COLMAP~\cite{schoenberger2016colmap} on the full resolution 3208x2200 images. Additionally, we compute one pointcloud for a neutral frame of each person, yielding 441 ground truth 3D geometries in total.

\newcolumntype{Y}{>{\centering\arraybackslash}X}
\newcolumntype{P}[1]{>{\centering\arraybackslash}p{#1}}
\begin{figure*}[htb!]
    \centering
    \includegraphics[width=0.99\textwidth]{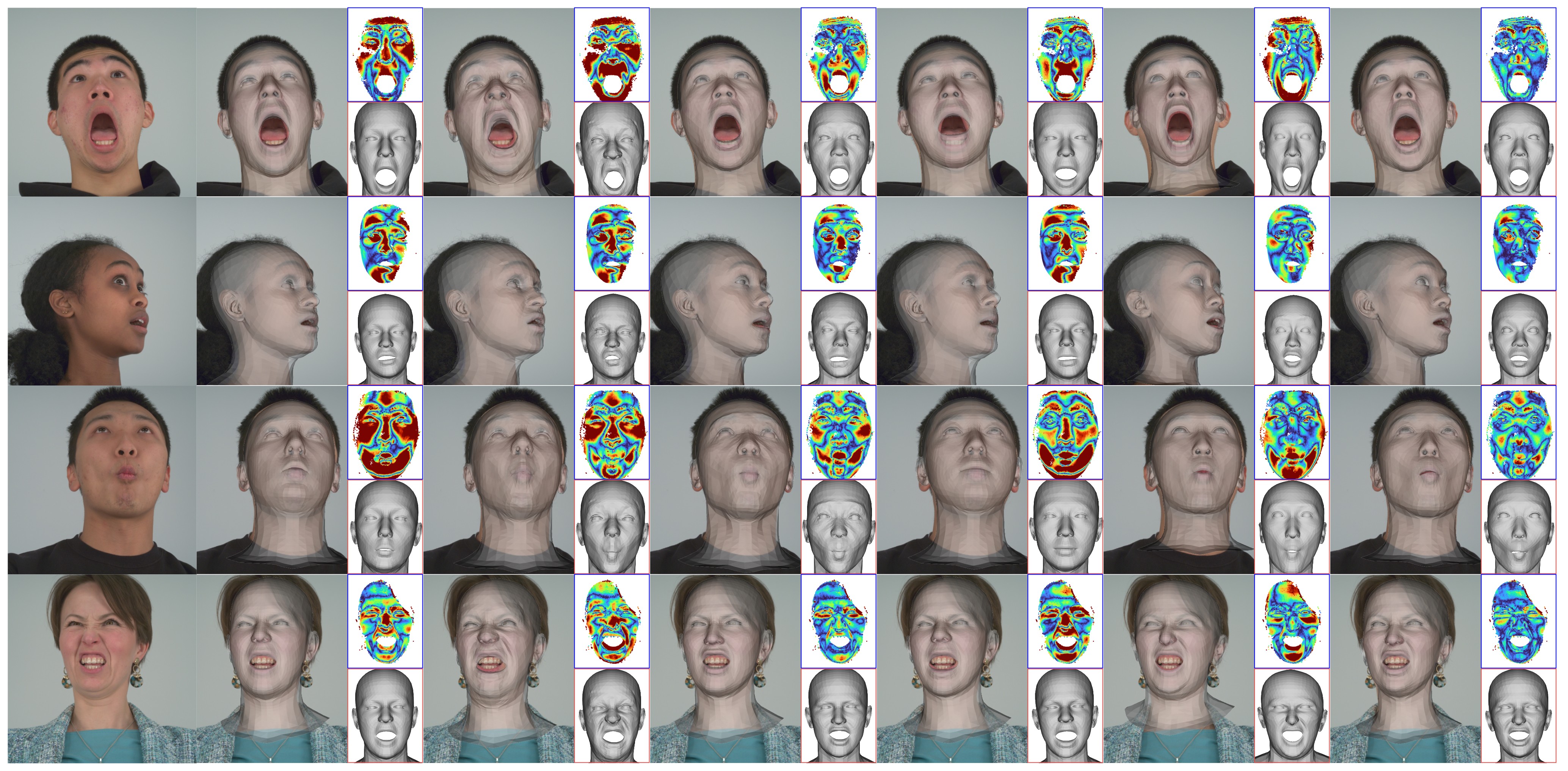}
    \setlength{\tabcolsep}{0pt}
    
    \begin{tabularx}{\textwidth}{P{0.12\linewidth}P{0.12\linewidth}YP{0.2\linewidth}YYY}
        \small{Input} & 
        \small{DECA}~\cite{feng2021learning} & 
        \small{EMOCA}~\cite{danvevcek2022emoca} &
        \small{Metrical Tracker}~\cite{zielonka2022towards} & 
        \small{TokenFace}~\cite{zhang2023accurate} & 
        \small{FlowFace}~\cite{taubner2024flowface} &
        \small{Ours}
    \end{tabularx}

    \caption{
    \textbf{Qualitative Comparison (Posed):} We show overlays of the reconstructed meshes to judge the reconstruction alignment. Insets with a blue border depict $L_2$-Chamfer distance as an error map, rendered from a frontal camera. Red insets show the reconstructed mesh from the same camera. We encourage the reviewers to watch our supplementary material for additional visualizations.
    }
    \label{fig:main_results}
\end{figure*}

\subsection{Task Description}
\label{sec:task_description}
Our benchmark consists of two 3D face reconstruction tasks: \textit{posed} and \textit{neutral} 3D face reconstruction. The posed reconstruction task aims to measure the fidelity of a 3D reconstruction. Given any expressive face image, the underlying geometry shall be recovered. This requires images with paired ground truth geometries which are available in NeRSemble trough COLMAP. The neutral reconstruction task on the other hand is specific to the face domain and measures how well a reconstruction method can disentangle the effects of shape and expression on a human 3D face. Specifically, the task is to reconstruct the geometry of a person's face under neutral expression given an image of the person under any arbitrary expression. 

\subsection{Evaluation Protocol}

To measure the performance of a reconstructed posed or neutral 3D face, we follow established practice and first rigidly align the prediction to the ground truth pointcloud via landmark correspondences and ICP. Furthermore, we use segmentation masks~\cite{zheng2022farl} to remove non-facial areas (hair, neck, ears, and mouth interior) from the ground truth. We then compute three metrics: (i) uni-directional Chamfer distance (L1 and L2) from GT points to the nearest mesh surface, (ii) cosine similarity (NC) of predicted mesh normals and GT pointcloud normals, (iii) Recall thresholded at 2.5mm (R\textsuperscript{2.5}) which is the percentage of GT points whose nearest mesh surface is 2.5mm or closer.

\newcolumntype{Y}{>{\centering\arraybackslash}X}
\newcolumntype{P}[1]{>{\centering\arraybackslash}p{#1}}
\begin{figure*}[htb!]
    \centering
    \includegraphics[width=0.99\textwidth]{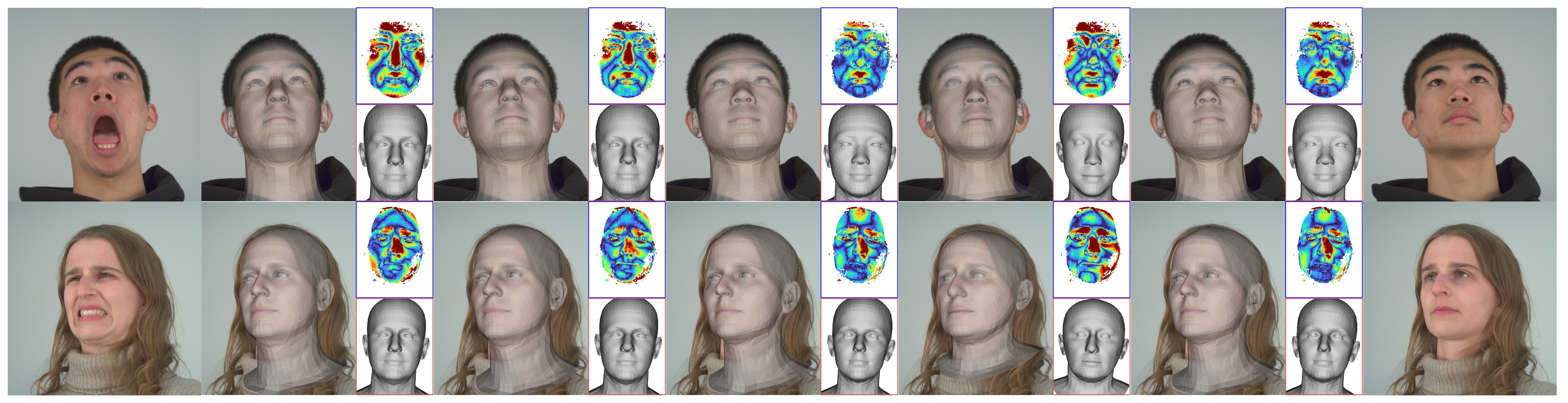}
    \setlength{\tabcolsep}{0pt}

    \begin{tabularx}{\linewidth}{YYYYYYY}
        \small{Input} & 
        \small{DECA}~\cite{feng2021learning} & 
        \small{EMOCA}~\cite{danvevcek2022emoca} &
        \small{MICA}~\cite{zielonka2022towards} & 
        \small{FlowFace}~\cite{taubner2024flowface} & 
        \small{Ours} &
        \small{Neutral image}
    \end{tabularx}

    \caption{
    \textbf{Qualitative Comparison (Neutral):} Alignment of the neutral prediction against the neutral image and scan of a person.
    }
    \label{fig:main_results_neutral}
\end{figure*}

\section{Experimental Results}
\label{sec:results}

\subsection{Implementation Details}

\paragraph{Prior Learning}
We train {\OURS} using the Adam~\cite{adam} optimizer, a batch size of $40$, and 2 A6000 GPUs, which takes 3 days until convergence. We use a learning rate of $1\times 10^{-4}$ for the prediction head and $1\times 10^{-5}$ for the DINO backbone. %
For simplicity we choose a light-weight network head. Using a DPT~\cite{ranftl2021dpt} head instead resolves the last remaining patch artifacts of the ViT-Base backbone but drastically increases runtime whithout improving down-stream reconstruction performance. 
Similarly, we find that replacing ViT-Base with Sapiens-300M~\cite{khirodkar2024sapiens} backbone (the smallest available Sapiens model) incurs high computational costs without reconstruction benefits.
We use 10\% of the subjects as validation set, and  exclude all the subjects from our benchmark from the training set.

\paragraph{FLAME Fitting}

We use the Adam optimizer with $\text{lr}_{\text{id}}\!=\! 0.001$  and $\text{lr}_{\text{ex}}\!=\!0.003$. We set $\lambda_{uv}\!=\!2000$, $\lambda_{n}\!=\!200$, $\lambda_{\text{id}}\!=\!0.15$ and $\lambda_{\text{ex}}\!=\!0.01$. We perform 500 optimization steps which takes 30 seconds in our unoptimized implementation.

\subsection{Baselines}

\paragraph{Feed-Forward FLAME Regressors}
The first category of approaches we compare against are feed-forward neural networks trained to predict FLAME parameters. We choose DECA~\cite{feng2021learning} and EMOCA~\cite{danvevcek2022emoca} as baselines which are trained in a self-supervised fashion on 2D data only. Additionally, we compare against MICA~\cite{zielonka2022towards}, which is trained solely on 3D data and only predicts identity parameters $\mathbf{z}_{\text{id}}$, and TokenFace~\cite{zhang2023accurate} which trained on a mixture of 2D and 3D data. Since TokenFace is not publicly available, the authors ran their method on the images that we provided.

\paragraph{Optimization-Based Approaches}
We compare against MetricalTracker~\cite{zielonka2022towards}, which optimizes against two sets of facial landmark predictions~\cite{bulat2017far, cao20133d} and a photometric term. Additionally, we compare against FlowFace~\cite{taubner2024flowface}, a recent method that predicts flow from the uv-space into image space, in order to predict 2D image-space vertex positions. Similar to \OURS, FlowFace also uses a dense 2D vertex loss, but predicts them in a quite different manner.
Note that all methods in this category rely on MICA estimates to initialize $\mathbf{z}_{\text{id}}$. Since FlowFace is not publicly available, as of yet, the authors ran their method on our benchmark images. In the future, we hope that our proposed benchmark will be adopted as a standard by the community to encourage further quantitative comparisons across methods.

\subsection{Our Benchmark}

\begin{table}[]
\centering
\setlength{\tabcolsep}{1.8pt}
\small
\begin{tabular}{lrrrrcrrrr}
\toprule
         & \multicolumn{4}{c}{Neutral}              && \multicolumn{4}{c}{Posed}                 \\
          \cmidrule(l){2-5}  \cmidrule(l){7-10}
         & L1\scriptsize{$\downarrow$}       & L2\scriptsize{$\downarrow$}        & NC\scriptsize{$\uparrow$}        & R\textsuperscript{2.5}\scriptsize{$\downarrow$}       &&  L1\scriptsize{$\downarrow$}       & L2\scriptsize{$\downarrow$}        & NC\scriptsize{$\uparrow$}        & R\textsuperscript{2.5}\scriptsize{$\downarrow$}       \\ \midrule
MICA~\cite{zielonka2022towards}     
    & 1.68 & 1.14 & \textbf{0.883} & 0.910 &&  -         &   -      &   -      &    -    \\
TokenFace~\cite{zhang2023accurate} 
    & - & - & - & - && 2.62 & 1.78 & 0.865 & 0.768 \\
DECA~\cite{feng2021learning}     
    & 2.07         &  1.40         &    0.876      &  0.845       &&  2.38         &   1.61        &   0.870      &     0.798    \\
EMOCAv2\cite{danvevcek2022emoca}  
    &    2.21      &   1.49        &    0.873      &    0.824     &&   2.63        &    1.78       &     0.860    &    0.758     \\
Metr. Tracker  
    & - & - & - & - &&    2.03 &  1.37 &  0.878 &  0.857      \\

FlowFace~\cite{taubner2024flowface} 
    & 1.93 & 1.31 &  0.878 &  0.870   && 1.96   & 1.33   & 0.879   & 0.879   \\
Ours     
    & \textbf{1.66}  & \textbf{1.12}  & \textbf{0.883}  & \textbf{0.912} && \textbf{1.66} & \textbf{1.11}   & \textbf{0.884} & \textbf{0.916} \\ \bottomrule
\end{tabular}
\caption{\textbf{Quantitative Comparison on Our Benchmark.}}
\label{tab:main_results}
\vspace{-0.3cm}
\end{table}

\paragraph{Posed Face Reconstruction}

We present quantitative and qualitative results for the posed reconstruction task (see \cref{sec:task_description}) in \cref{tab:main_results} and \cref{fig:main_results}, respectively. 
Quantitatively, {\OURS} outperforms all baselines by a large margin. 
In general, the feed-forward predictors (DECA, EMOCAv2, TokenFace) perform significantly worse than the optimization based approaches (MetricalTracker, FlowFace and Ours). Visually, DECA and TokenFace seem to underfit facial expressions, while EMOCAv2 exaggerates them. 
Compared to our approach, FlowFace sometimes exhibits performance drops for extreme facial expressions.

\newcolumntype{Y}{>{\centering\arraybackslash}X}
\newcolumntype{P}[1]{>{\centering\arraybackslash}p{#1}}
\begin{figure*}[htb!]
    \centering
    \includegraphics[width=0.99\textwidth]{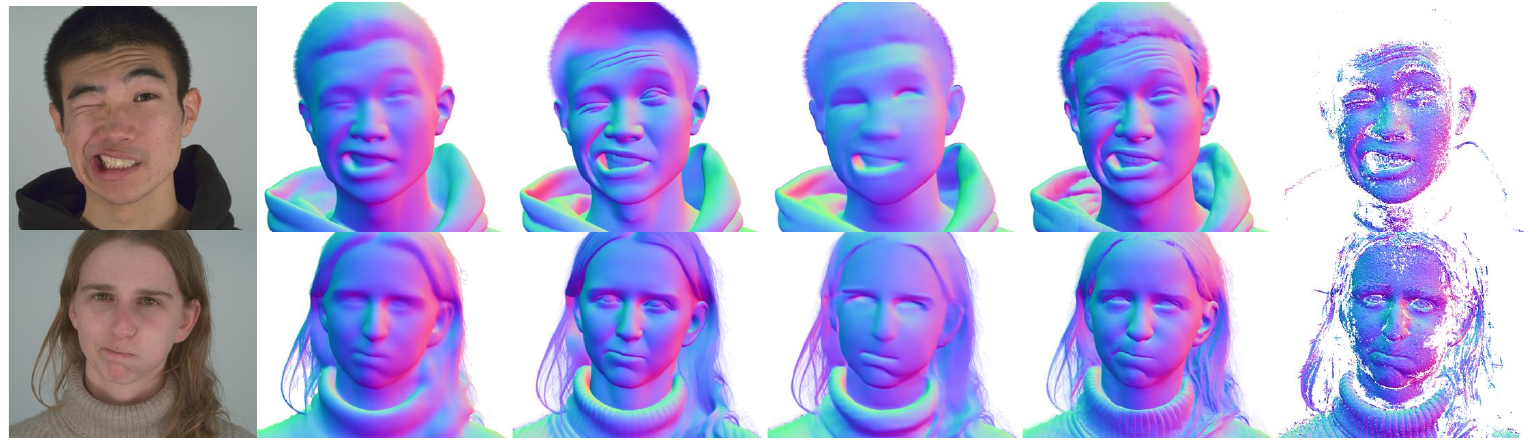}

     \begin{tabularx}{\textwidth}{
         YYYYYY
     }
    \small{Input} & 
    \small{Sapiens-2B} & 
    \small{Diff-E2E} &
    \small{Metric3D} & 
    \small{Ours} & 
    \small{COLMAP}
    \end{tabularx}
    \caption{
    \textbf{Surface Normal Estimation:} Qualitative comparison to state-of-the-art surface normal estimators. From left to right we show the single input image and the predictions of Metric3D~\cite{hu2024metric3d}, Sapiens-2B~\cite{khirodkar2024sapiens}, Diff-E2E~\cite{martingarcia2024diffusione2eft}, our result and COLMAP~\cite{schoenberger2016colmap} normals.
    }
    \label{fig:normals}

\end{figure*}

\paragraph{Neutral Face Reconstruction}

Results on the neutral reconstruction task (see \cref{sec:task_description}) are provided in \cref{fig:main_results_neutral} and \cref{tab:main_results}.
First of all, we can observe that the significantly better posed reconstruction metrics of FlowFace and {\OURS} do not immediately translate to the neutral reconstruction. We attribute this to the ambiguities between identity and expression in the optimization process. Note that both FlowFace and {\OURS} rely on MICA predictions to initialize identity parameters $\mathbf{z}_{\text{id}}$. While FlowFace ends up with worse neutral reconstructions, our approach is able to improve upon MICA by a small margin. Nevertheless, we highlight the importance of using MICA to help disambiguate between $\mathbf{z}_{\text{id}}$ and $\mathbf{z}_{\text{ex}}$, as ablated in \cref{sec:ablations}. Note, that TokenFace is missing from the neutral evaluation, since the authors only provided posed meshes.

\subsection{Results on Existing Benchmarks}

\paragraph{FaceScape Benchmark~\cite{zhu2023facescape}}
The FaceScape benchmark only evaluates the posed reconstruction task. The relative performance across methods matches with results on our benchmark, see \cref{tab:now_and_facescape}. Our method outperforms all baselines by a large margin w.r.t. chamfer distance (CD) and mean normal error (MNE), and has a slightly worse completeness rate (CR) than DECA, see \cite{zhu2023facescape} for more details.

\begin{table}[]
\centering
\small
\setlength{\tabcolsep}{2.5pt}
\begin{tabular}{lrrrcrrr}
\toprule
\multirow{2}{*}{Method} & \multicolumn{3}{c}{NoW~\cite{Sanyal2019now}}        && \multicolumn{3}{c}{FaceScape~\cite{zhu2023facescape}} \\ 
                            \cmidrule(l){2-4} \cmidrule(l){6-8} 
                        & Median$\downarrow$     & Mean$\downarrow$    & Std$\downarrow$     && CD$\downarrow$   & MNE$\downarrow$   & CR$\uparrow$\\ 
                \midrule
Dense~\cite{wood2022denselandmarks}     & 1.02       & 1.28    & 1.08    && -        & -         & - \\
PRNet~\cite{wang2019prnet}              & -          & -       & -       && 3.56     & 0.126     & 0.896 \\
3DDFAv2~\cite{guo20203ddfav2}           & -          & -       & -       && 3.60     & 0.096     & 0.931            \\
DECA~\cite{feng2021learning}            & 1.09       & 1.38    & 1.18    && 4.69     & 0.108     & \textbf{0.995}  \\
MICA~\cite{zielonka2022towards}         & 0.90       & 1.11    & 0.92    && -        & -         & - \\
FlowFace~\cite{taubner2024flowface}     & 0.87       & 1.07    & 0.88    && 2.21     & 0.083     & -  \\
TokenFace~\cite{zhang2023accurate}      & \textbf{0.76}  & \textbf{0.95} & \textbf{0.82} && 3.70 & 0.101 & 0.938    \\
Ours                                    & 0.87       & 1.07    & 0.89      && \textbf{1.76}    & \textbf{0.077}      &  0.980 \\ 
                \bottomrule
\end{tabular}
\caption{\textbf{NoW}~\cite{Sanyal2019now} \textbf{and FaceScape}~\cite{zhu2023facescape}\textbf{ Benchmark}.}
\label{tab:now_and_facescape}

\end{table}

\paragraph{NoW Benchmark~\cite{Sanyal2019now}}

On the NoW benchmark, which only evaluates the neutral reconstruction task, we achieve the same metrics as FlowFace, which is the best-performing optimization based approach, but perform worse than TokenFace. Note, however, that on FaceScape and our benchmark, we significantly outperform TokenFace. Similarly to the results on our benchmark, {\OURS} can only improve a small amount on top of the MICA predictions. We hypothesize that our prior significantly helps posed reconstructions, but struggles to guide the optimization to properly disentangle between $\mathbf{z}_{\text{id}}$ and $\mathbf{z}_{\text{ex}}$.

\subsection{In-the-Wild Results}
In \cref{fig:teaser}, we demonstrate the robustness of our prior networks and fitting algorithm on challenging in-the-wild examples, including strong appearance variation, various background contexts and surroundings, lighting/shadow effects, and occlusions such as glasses, head wear and hands.
Ultimately, this demonstrates that our approach successfully generalizes, even beyond the training data distribution. We hope that this will inspire more work in a similar direction, especially since all data is available and 2 48GB GPUs are sufficient for training.\\
For tracking results on in-the-wild monocular videos we refer the reader to our supplementary video.

\subsection{Surface Normal Estimation}

\begin{table}[]
\small
\centering
\setlength{\tabcolsep}{2.4pt}
\begin{tabular}{lcccc}
\toprule
    \multicolumn{2}{r}{\hspace{1.5cm}Metric3D\cite{hu2024metric3d}} & Sapiens-2B\cite{khirodkar2024sapiens} & Diff-E2E\cite{martingarcia2024diffusione2eft} & Ours \\
\midrule
    Normal Sim.$\uparrow$
        & 0.900 & 0.913 & 0.913 & \textbf{0.931} \\
\bottomrule

\end{tabular}
\caption{\textbf{Normal Estimation}: We report the cosine similarity of predicted normals against 16-view COLMAP~\cite{schoenberger2016colmap} estimates. The results are averaged over all images from our benchmark.}
\label{tab:normals}

\end{table}
In \cref{tab:normals} and \cref{fig:normals}, we show quantitative and qualitative comparisons against recent state-of-the-art normal estimation methods~\cite{khirodkar2024sapiens,martingarcia2024diffusione2eft,hu2024metric3d}. Our network estimates more detailed and accurate normals than the baselines.

\subsection{Ablation Experiments}

\begin{table}[]
\setlength{\tabcolsep}{4.3pt}
\small
\centering
\begin{tabular}{lrrrcrrr}
\toprule
    & \multicolumn{3}{c}{Neutral}               && \multicolumn{3}{c}{Posed}                   \\ 
       \cmidrule(l){2-4}                           \cmidrule(l){6-8}
    & L1\scriptsize{$\downarrow$}       & L2\scriptsize{$\downarrow$} & R\textsuperscript{2.5}\scriptsize{$\downarrow$}       && L1\scriptsize{$\downarrow$}       & L2\scriptsize{$\downarrow$} & R\textsuperscript{2.5}\scriptsize{$\downarrow$}       \\
\midrule
Lmks. 
    & 1.68 &  1.14 &  0.911 && 2.02 &  1.37 &  0.857 \\
Lmks. + Pho.        
    & 1.69 &  1.14 &  0.908 && 2.05 &  1.38 &  0.854 \\

Ours, only $\mathcal{U}$
    & \textbf{1.66}  & \textbf{1.11} & \textbf{0.913} && 1.72 &   1.16   & 0.906 \\
Ours, only $\mathcal{N}$      
    & 1.69  & 1.12  & 0.907 && 1.70 & 1.14 & 0.910 \\
Ours, only Sapiens     
    & 1.72  & 1.16  & 0.902 && 1.81 & 1.23 & 0.890 \\
\midrule
Ours            
    & \textbf{1.66}  & 1.12  & 0.912 && \textbf{1.66} & \textbf{1.11}   & \textbf{0.916} \\ 
    \midrule
    Ours, no MICA
    & 1.90 & 1.29 &  0.872 && 1.74 &  1.17  &  0.901 \\
    \bottomrule
\end{tabular}
\caption{\textbf{Fitting Algorithm Ablations:} We compare different compositions of our optimization energy $E$, see \cref{eq:energy}.
}
\label{tab:ablations}

\end{table}
\label{sec:ablations}

We conduct extensive ablations on different compositions of our optimization energy $E$ in \cref{tab:ablations}. We start by using the simplest energy, with only the landmark loss from MetricalTracker, and our regularization term. Next we add a photometric term, as in MetricalTracker. As shown in \cref{tab:ablations}, these configurations achieve significantly worse posed reconstructions.
Next, we investigate the effect of only using the predictions from $\mathcal{N}$ and $\mathcal{U}$, respectively. Compared to our full model these variants showcase lower posed reconstruction scores. We also compare our normal predictor $\mathcal{N}$ against Sapiens-2B~\cite{khirodkar2024sapiens}, which confirms that our improved normal predictions translate to better reconstructions.
Finally, we ablate the effect of using the MICA prior. Without MICA's predictions of $\mathbf{z}_{\text{id}}$ especially the neutral reconstruction metrics drop, indicating its importance for disentanglement between identity and expression.

\section{Limitations and Future Work}
\label{sec:limitations}

While we demonstrate the effectiveness of our approach for single image 3D reconstruction, several limitations remain.
While our optimization energy could be easily extended to incorporate  observations from multiple viewpoints, our prior models cannot  currently exploit multiview information. Future extensions of our architecture could include multiview inputs similar to DUSt3R~\cite{wang2024dust3r}, or video inputs similar to RollingDepth~\cite{ke2024rollingdepth}.
Next, for training large-scale 3DMM conditioned generative models like 3D GANS~\cite{sun2023next3d} or diffusion models~\cite{kirschstein2024diffusionavatars,prinzler2024joker,taubner2024cap4d}, e.g. on the LAION-Face dataset~\cite{zheng2022farl}, fast reconstruction speed would be desirable. One potential avenue could be the distillation of our per-pixel predictors into a feed-forward 3DMM predictor.
Finally, our experiments showcase, that optimization based approaches cannot flawlessly disambiguate identity and expression parameters. Therefore, specifically crafted priors for disambiguation are required.

\section{Conclusion}
\label{sec:conclusion}
In this paper, we trained pixel-aligned geometric prior networks, by leveraging pre-trained, generalized foundational features on publicly available 3D face datasets, which we registered into a uniform format. 
Our trained networks successfully generalize beyond the diversity of the training data, and we experimentally show that our normal predictor significantly outperforms all available normal estimators.
We designed a 3DMM fitting algorithm on top of our prior predictions, which results in state of the art single image 3D reconstruction. 
Finally, we introduce a new benchmark, which features diverse and extreme expressions and allows, for the first time, to simultaneously evaluate neutral and posed geometry.

\section*{Acknowledgements}
This work was funded by Synthesia and supported by the ERC Consolidator Grant Gen3D (101171131), the German Research Foundation (DFG) Research Unit ``Learning and Simulation in Visual Computing''. Additionally, we would like to thank Angela Dai for the video voice-over.

{
    \small
    \bibliographystyle{ieeenat_fullname}
    \bibliography{main}
}

\end{document}